**Accuracy and precision: a new view on kinematic assessment of solid-state hinges and compliant mechanisms**

LF Campanile, S Kirmse and A Hasse (corresponding author),
E-Mail: alexander.hasse@mb.tu-chemnitz.de
Chemnitz University of Technology, Professorship Machine Elements and Product Development, Chemnitz, Germany

**Abstract**

Compliant mechanisms are alternatives to conventional mechanisms which exploit elastic strain to produce desired deformations instead of using moveable parts. They are designed for a kinematic task (providing desired deformations) but do not possess a kinematics in the strict sense. This leads to difficulties while assessing the quality of a compliant mechanism's design.

The kinematics of a compliant mechanism can be seen as a fuzzy property. There is no unique kinematics, since every deformation need a particular force system to act; however, certain deformations are easier to obtain than others. A parallel can be made with measurement theory: the measured value of a quantity is not unique, but exists as statistic distribution of measures. A representative measure of this distribution can be chosen to evaluate how far the measures divert from a reference value. Based on this analogy, the concept of accuracy and precision of compliant systems are introduced and discussed in this paper.

A quantitative determination of these qualities based on the eigenvalue analysis of the hinge's stiffness is proposed. This new approach is capable of removing most of the ambiguities included in the state-of-the-art assessment criteria (usually based on the concepts of path deviation and parasitic motion).

**Introduction**

Compliant mechanisms are alternatives to conventional mechanisms which exploit elastic deformations instead of finite motions at discontinuities like hinges or guides. Solid-state hinges are a special case of compliant mechanisms. They join two rigid bodies (links) while partially restraining their relative motion. For the sake of brevity and clarity, we will put a special focus on solid-state hinges in this work; however, all presented concepts and findings can be easily extended to compliant mechanisms in general. In the following, the term "solid-state hinge" will include, besides revolute hinges (designed for rotational relative motion between the links), also guides (designed for relative translation). Conventional as well as compliant mechanisms are designed to restrain their deformation to a number $p$ of degrees of freedom, which is called *mobility* in the conventional case and *pseudo-mobility* (Hasse and Campanile, 2009) in the compliant case. The first part of our considerations will be focused on the case $p$=1.

Conventional mechanisms are defined by their kinematics. Their layout determines which deformations are possible and which ones are impossible, and applied loads have no influence on this discrimination. Compliant mechanisms, instead, are elastic bodies and therefore defined by a functional relationship between loads and displacements. Their deformations are load-dependent and so they possess no kinematics (for an elastic body, all deformation are possible when it is loaded accordingly). Nevertheless, most compliant mechanism are (implicitly or explicitly) asked to mimic a conventional mechanism or realize a given kinematics, and so some kind of kinematic assessment of compliant mechanisms and solid-state hinges is needed.

In the case of solid-state hinges with $p$=1 and one link fixed, kinematic considerations can be reduced to the study of the paths described by the points of the moving link. In a conventional hinge, all points of the moving link follow fixed paths, i.e. the paths do not depend on the loading. In solid-state hinges, instead, the paths change with the loading. If a solid-state hinge is to be compared with a conventional hinge, the question whether they are kinematically equivalent or by which extent they kinematically divert from another cannot be answered in a strict sense, since this would imply comparing a load-dependent path with a fixed one. A structured answer is needed, which covers two aspects: the first one concerns how load-sensitive the paths are; the second one deals with the comparison between the kinematics of the solid-state hinge (which only exists in a fuzzy sense) with the kinematics of the conventional hinge. This suggest an analogy with measurement theory. There is nothing such a unique measured value, but different values scattered over a certain range. Their overall closeness to a reference value ("true value") defines the *accuracy* of a measure, while the mutual closeness of the single measurements expresses the measurement's *precision*. We will therefore denote the closeness of the hinge's kinematics to a given reference as accuracy and its robustness against applied loads as precision of the hinge.

As in the measurement case, a quantitative determination of the accuracy requires the election of a representative (among the paths under loading) by some sort of convention.

Precision is an inherent quality of the solid-state hinge, while accuracy depends on a specified reference. The difference between the real path of the hinge and the one of a given reference can so be split in an accuracy-related deviation, which is loading-independent, and a precision-related deviation, which depends on the applied load.

Generally, a linear guide is designed to realize a rectilinear translation of the moving link. This desired motion can be therefore taken as reference kinematics for a compliant linear guide like the one in Figure 1 (*reference path*). In reality, the link performs a translation along a curvilinear path (*real path*). Depending on the load, the real path of the guide's link changes. According to a chosen convention, one particular real path can be chosen to represent the real mechanism's kinematics (*representative path*). The guide's accuracy is defined with respect to the deviation between representative and reference path (*path deviation*). The path variation under load, i.e. the scattering of the real paths, describes the guide's precision.

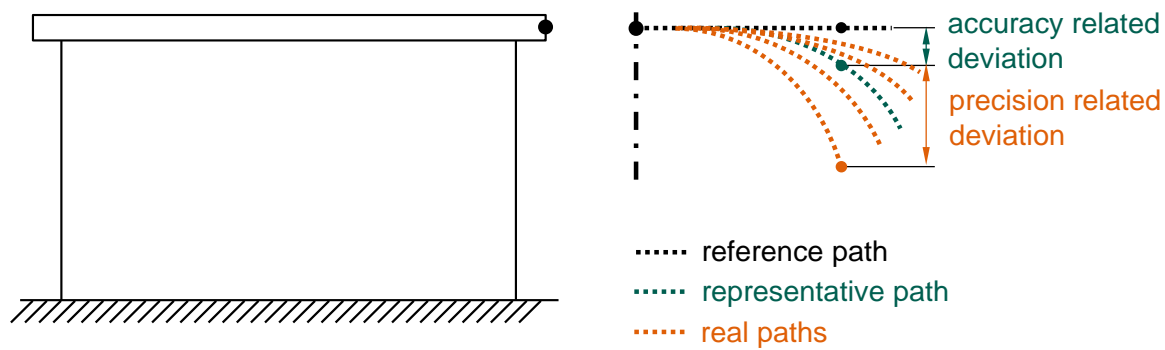

Figure 1: Accuracy and precision of a compliant linear guide

**Kinematic studies on solid-state hinges in literature**

The question of kinematic assessment of solid-state hinges is not covered in a unique and consistent way in the literature. An approach like the proposed one, which takes into account the fuzzy nature of kinematics in compliant mechanism, cannot be found in previous studies.

A first issue in published work on kinematic analysis of solid-state hinges consists in the ambiguous use of terms: the concept of *parasitic error* is interchangeably used in the accuracy sense (i.e. as a deviation from a reference), as well as to denote the response of the hinge to transverse loads, which is as mentioned before – affected by both precision and accuracy. Sometimes the term "parasitic motion" is used instead of "parasitic error". The term "path deviation" is, on the other hand, consequently used in the accuracy sense. For revolute hinges, the concept of axis drift is also used, which expresses the displacement of the instantaneous axis of rotation of the relative motion between the hinge's links. The axis drift is directly related to the path deviation when the kinematics of a conventional pivot joint is taken as a reference.

Analyses of the axis drift can be found for example in (Goldfarb and Speich, 1999; Fowler et al., 2014; Xu and King, 1996; Linß et al., 2017; Du et al., 2020; Marković and Zelenika, 2017). The axis drift is recorded or calculated as a function of the hinge swing angle. However, the pivot displacement is only determined for loading by a single force, whose point of application is often not provided (Goldfarb and Speich, 1999; Fowler et al., 2014; Du et al., 2020; Marković and Zelenika, 2017). No attempt is made to define a representative path; consequently, the path deviation is not uniquely defined. Even where the loading is specified, it represents just one arbitrary choice, so that the resulting path deviation cannot be taken (without additional evidence) as a representative measure of the hinge's accuracy.

Concerning the response to transverse loads, several contributions deal with the determination of *stiffnesses* (e.g. in (Goldfarb and Speich, 1999; Fowler et al., 2014; Trease et al., 2005)) or *compliances* (Lobontiu and Paine, 2002; Lobontiu et al, 2001; Smith et al., 1997; Lobontiu et al., 2013; Weisbord und Paros, 1965; Bereslli et al. 2013) related to certain degrees of freedom defined a-priori as parasitic. This arbitrary choice represents the central critical aspect of these analyses. They do not look for an inherent property of the hinge to express how much its kinematics is robust against external loads in general, but prescribes a-priori which kinematics the hinge is expected to follow and studies the deviations from this kinematics under load.

As mentioned, however, the hinge's precision does not depend on an external, a-priori chosen reference. It is an inherent property of the hinge. If a conventional revolute hinge would be loaded to determine its precision, any load component acting along its kinematics (any circumferential component) would generate indefinitely large displacements. They are not relevant, however, since they would not divert from the system's kinematics. In order to properly assess the hinge's precision, hence, the knowledge of the hinge's kinematics is needed, either to select the forces beforehand (and apply only transverse loads – in this case radial forces) or to properly filter the response. Compliant mechanisms do not have a kinematics in the "sharp" sense, but they possess one in the fuzzy sense, which needs to be taken into account. If a solid-state hinge is loaded along a "preferred" path (i.e. a path which it "likely" follows) deformations will be less relevant for the hinge's precision than if the loads act transversely to this path. A pure precision assessment therefore uses (implicitly or explicitly) an *inherent kinematics* as a representative for the system's (fuzzy) kinematics and serves as reference for the definition of transverse loads and parasitic displacements. Such an approach is not present among the state-of-the-art criteria.

Additionally, the sensitivity of hinges towards transverse loads is always assessed, in published work, with respect to particular load cases. A global, load-case independent quantity is not provided.

Finally, the determination of parasitic displacements or properties like stiffness and compliance provides dimensional figures, which are not meaningful if they are not related to the quantitative data about size and elasticity modulus of the hinge.

The inherent kinematics mentioned above can be used to fill the other main gap left by the state-of-the-art studies, i.e. the absence of a qualified, standardized reference to be used in the accuracy assessment.

**Definition of accuracy and precision based on the stiffness eigenproperties**

In the linearized case (small rotations), a proper basis for the definition of the missing standard is given by the eigenvalue analysis of the hinge's stiffness, as will be shown in detail.

An extension to the general case of large rotations is conceivable on the basis of a local linearization and on the eigenvalue analysis of the tangent stiffness matrix. Such option will not be treated here and will be the object of future research.

The relationship between the eigensystem of the stiffness matrix of a solid-state hinge or a compliant mechanism and the respective kinematic behaviour was already addressed in a design context. In (Campanile et al., 2004) a modal procedure for the design of compliant mechanisms for shape adaptation was presented. The procedure aims at assigning a given deformation distribution as the first eigenvector of the stiffness matrix and manipulating the eigenvalues such that the assigned deformation distribution dominates the static response. The modal procedure was generalized and combined with formal optimisation in (Hasse and Campanile, 2008) and further developed in later papers (Hasse and Campanile, 2009; Hasse et al., 2011; Hasse, 2016).

In the linear case, for a conventional hinge with mobility equal to one, the displacement field of the relative motion can be expressed as the product of a fixed spatial function (displacement distribution) and a variable, scalar value (displacement amplitude). The fixed displacement distribution rules the possible static responses and therefore represents the hinge's kinematics. From an energetic point of view, it is infinitely more convenient: it costs zero energy, while any other displacement distribution involves a virtually infinite amount of energy.

In search of the inherent kinematics of a solid-state hinge, we can therefore look for displacement distributions with a low level of strain energy.

If the deformation behaviour of the hinge (e.g. after a FEM discretization) is described by the equation

$$\mathbf{k}\mathbf{u} = \mathbf{f} \quad (1)$$

with $\mathbf{u}$ as vector of the displacements, $\mathbf{f}$ as vector of the corresponding forces and $\mathbf{k}$ as stiffness matrix, the expression of the strain energy is

$$E = \frac{1}{2}\mathbf{u}^T \mathbf{k} \mathbf{u} \quad (2)$$

No unique value of the energy can be assigned to a kinematics, i.e. to a displacement distribution. If a displacement field $\mathbf{u}$ is scaled by a factor $\alpha$, the corresponding energy is scaled by $\alpha^2$ with no change in the displacement distribution. The problem can be solved by a suitable normalization. If the normalization criterion is selected

$$\mathbf{u}^T\mathbf{u} = \frac{1}{2} \tag{3}$$

the Rayleigh quotient is assigned to each displacement field as a measure of the energy:

$$R = \frac{\mathbf{u}^T\mathbf{k}\mathbf{u}}{\mathbf{u}^T\mathbf{u}} \tag{4}$$

Note that the right side of the equation (3) is not a pure number, but has the dimension of the square of a length. As intended, the Rayleigh quotient as a measure of the energy of a displacement distribution is invariant with respect to a scaling of the displacement field.

The inherent kinematics of the system can now be chosen as the displacement distribution which minimizes the Rayleigh quotient. According to the Courant-Fischer theorem, this displacement distribution corresponds to the first eigenvector $\boldsymbol{\varphi}_1$ of the stiffness matrix. The (minimum) value of the Rayleigh quotient is then given by the first eigenvalue $\lambda_1$:

$$\min_{\mathbf{u} \neq \mathbf{0}} \frac{\mathbf{u}^T\mathbf{k}\mathbf{u}}{\mathbf{u}^T\mathbf{u}} = \frac{\boldsymbol{\varphi}_1^T\mathbf{k}\boldsymbol{\varphi}_1}{\boldsymbol{\varphi}_1^T\boldsymbol{\varphi}_1} = \lambda_1 \tag{5}$$

The eigenvalues $\lambda_i, i=1...n$ (sorted in ascending order) and the eigenvectors $\boldsymbol{\varphi}_i, i=1...n$ of the matrix $\mathbf{k}$ are determined by solving the eigenvalue problem

$$\mathbf{k}\boldsymbol{\varphi} = \lambda\boldsymbol{\varphi} \tag{6}$$

The number $n$ corresponds to the number of degrees of freedom of the system. The eigenvectors can be normalized according to the criterion

$$\boldsymbol{\varphi}^T\boldsymbol{\varphi} = 1 \tag{7}$$

Through the modal transformation:

$$\mathbf{u} = \sum_i a_i \boldsymbol{\varphi}_i \tag{8}$$

and the definition of the modal forces

$$q_i = \boldsymbol{\varphi}_i^T \mathbf{f} \tag{9}$$

the decoupled system equations are obtained

$$\lambda_i a_i = q_i \tag{10}$$

According to (10), $n$ springs exist in the modal space whose static deflections represent the modal amplitudes. The corresponding values of the spring stiffness are the eigenvalues (see Figure 2).

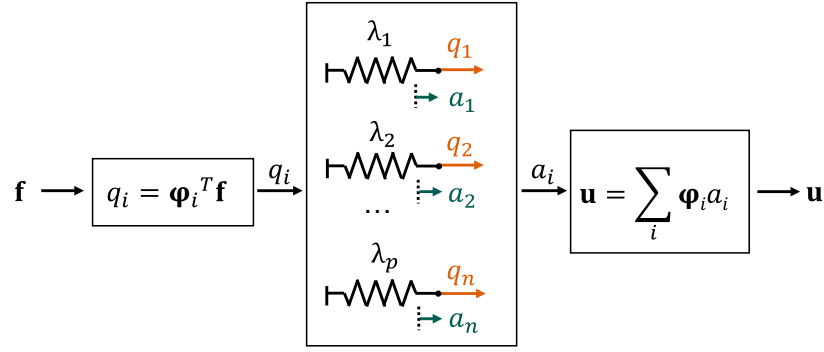

Figure 2: Deformation behaviour in modal space

In general, it can be expected that the first eigenvector has a relatively large share in the system response, since $\lambda_1$ is the smallest eigenvalue and therefore the first modal coordinate

$$a_1 = \frac{q_1}{\lambda_1} \tag{11}$$

will tend to be larger than the others. This behaviour will be more pronounced the higher the selectivity (Hasse and Campanile, 2009)

$$S = \frac{\lambda_2}{\lambda_1} \tag{12}$$

is. For a given load case (i.e. for given $\mathbf{q}$), the deviation of the response from the first eigenvector becomes smaller as $S$ increases. It can also be assumed that the number of load cases where the deviation falls within a given tolerance increases with increasing $S$.

The first eigenvector $\boldsymbol{\varphi}_1$ hence represents the searched inherent kinematics of the hinge, which will be called *natural kinematics*. The selectivity $S$ provides a suitable precision measure. As desired, it describes the sensitivity of the hinge kinematics to transverse loads without the need of an external, arbitrary reference. Choosing the selectivity as precision measure also satisfies the needs of a load-case independent assessment and of a dimensionless, size-independent figure.

If one considers the natural kinematics of the hinge as desired and the remaining components as parasitic, the selectivity represents the order of magnitude of the ratio between the displacements related to the desired deformation and those of the parasitic components.

The natural kinematics can be used to assess the hinge's accuracy unambiguously. After having defined the reference kinematics as vector $\boldsymbol{\varphi}_r$ of the displacement space, subject to the normalization condition (8), its deviation from $\boldsymbol{\varphi}_1$ will provide the searched accuracy measure. The comparison between the two vectors can be made according to the cosine similarity

$$\delta = \left|\cos(\vartheta)\right| = \left|\boldsymbol{\varphi}_1^T \boldsymbol{\varphi}_r\right| \tag{13}$$

with $\vartheta$ as angle between the vectors. The accuracy index $\delta$ is equal to one in the case of perfect accuracy and zero when the vectors are orthogonal to another.

The presented concepts of natural kinematics, selectivity as precision measure and accuracy index are of general validity and can also be applied to compliant mechanisms without any modification.

The above mentioned accuracy measures (path deviation and axis drift) can be computed (when defined), from the natural and reference kinematics, with procedures which are case-specific.

For solid-state hinges and compliant mechanisms with pseudo-mobility $p$ larger than one, light modifications are necessary. The natural and the reference kinematics will be represented by $p$ vectors and the selectivity will turn to

$$S = \frac{\lambda_{p+1}}{\lambda_p} \tag{14}$$

The reference kinematics is defined by a set of $p$ normalized vectors

$$\Phi_r = \begin{bmatrix} \varphi_{r1} & | & \varphi_{r2} & | & \cdots & | & \varphi_{rp} \end{bmatrix} \tag{15}$$

The subspace of $\mathbb{R}^n$ spanned by $\Phi_r$ is to be compared with the subspace spanned by the first $p$ normalized eigenvectors of the stiffness matrix

$$\Phi_e = \begin{bmatrix} \varphi_1 & | & \varphi_2 & | & \cdots & | & \varphi_p \end{bmatrix} \tag{16}$$

This is done by means of the *extended cosine similarity* (Campanile et al., 2021)

$$\delta_e = \sqrt{\beta_1} \tag{17}$$

where $\beta_1$ is the smallest eigenvalue of the problem

$$\Phi_r^T \Phi_e \Phi_e^T \Phi_r \mathbf{b} = \beta \mathbf{b} \tag{18}$$

**Precision measure -Application to a circular notch hinge**

To illustrate the introduced concepts, a so-called circular notch hinge with the dimensions shown in Figure 3 was chosen as an example. The thickness $l$ of the hinge's flexible area was varied. The first eigenmode describes a bending deformation in the desired direction and the second eigenmode describes a transverse bending of the hinge.

It can be seen that the selectivity as ratio of the first two eigenvalues depends on $l$. It decreases with increasing $l$.

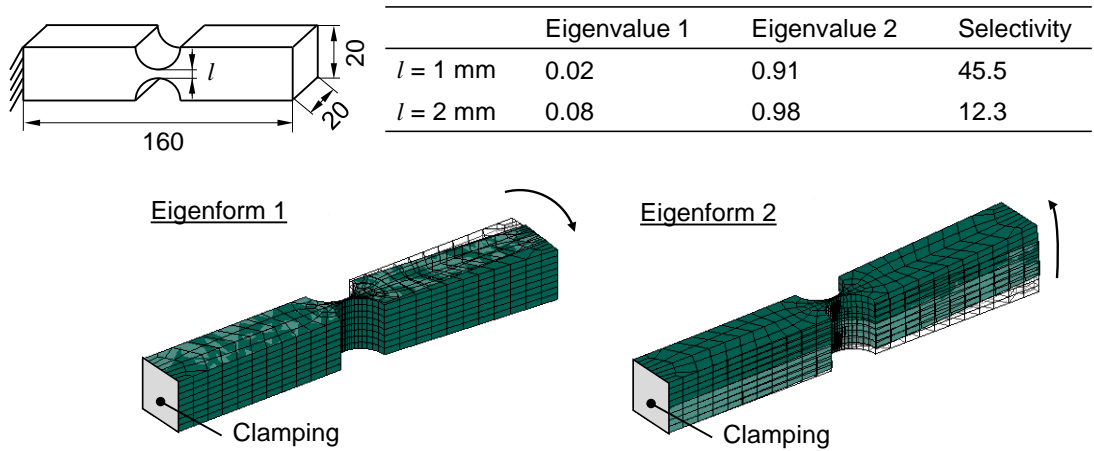

Figure 3: Eigenvalues and selectivities of the circular notch hinge with selected geometries.

**Outlook**

This paper proposes a standard for the kinematic assessment of compliant mechanisms and specifies it for the case of solid-state hinges designed for one kinematic degree of freedom (pseudo-mobility equal to one). The standard focuses on the concepts of accuracy and precision and is based on the study of the stiffness matrix's eigensystem. The first eigenvector of the stiffness matrix, defined as natural kinematics, is used to split the deviation of the hinge's kinematics from a reference into a precision-related, loading-dependent and reference-independent part, and an accuracy-related, load-independent and reference-dependent part. This standard can contribute to eliminate uncertainties and compensate omissions present in previous evaluation criteria and to simplify the comparison among designs of compliant mechanisms and corresponding design methods.

**Funding**

This work was funded by the Deutsche Forschungsgemeinschaft (DFG, German Research Foundation) – project number HA 7893/3-1.

**References**

Berselli G, Rad FP, Vertechy R and Castelli VP (2013) Comparative evaluation of straight and curved beam flexures for selectively compliant mechanisms. In *2013 IEEE/ASME International Conference on Advanced Intelligent Mechatronics* (pp. 1761-1766). IEEE.

Campanile LF, Kirmse S and Hasse A (2021) A measure for the similarity of vector spaces. *Researchgate*, Preprint, viewed 16 April 2021, DOI: 10.13140/RG.2.2.30164.73606

Campanile LF, Rose M and Breitbach EJ (2004) Synthesis of flexible mechanisms for airfoil shape control: a modal procedure, *Proc. of the 15th International Conference on Adaptive Structures and Technologies*, Bar Harbour, ME, USA.

Du S, Liu J, Bu H and Zhang L (2020) A novel design of a high-performance flexure hinge with reverse parallel connection multiple-cross-springs. *Review of Scientific Instruments*, *91*(3), 035121.

Fowler RM, Maselli A, Pluimers P, Magleby SP and Howell LL (2014) Flex-16: A large-displacement monolithic compliant rotational hinge. *Mechanism and Machine Theory*, *82*, 203-217.

Goldfarb M, and Speich JE (1999) A Well-Behaved Revolute Flexure Joint for Compliant Mechanism. *Journal of Mechanical Design,* 121(3): 424–429.

Hasse A (2016) Topology Optimization of Compliant Mechanisms Explicitly Considering Desired Kinematics and Stiffness Constraints. In: *International Design Engineering Technical Conferences and Computers and Information in Engineering Conference* (Vol. 50152, p. V05AT07A029). American Society of Mechanical Engineers.

Hasse A and Campanile F (2008) Synthesis of Compliant Mechanisms for Shape Adaptation by means of a Modal Procedure. In *19th International Conference on Adaptive Structures and Technologies (ICAST).*

Hasse A, and Campanile LF (2009) Design of compliant mechanisms with selective compliance. *Smart Materials and Structures*, *18*(11), 115016.

Hasse A, Zuest I and Campanile LF (2011) Modal synthesis of belt-rib structures. *Proceedings of the Institution of Mechanical Engineers, Part C: Journal of Mechanical Engineering Science*, *225*(3), 722-732.Linß S, Schorr P and Zentner L (2017) General design equations for the rotational stiffness, maximal angular deflection and rotational precision of various notch flexure hinges. *Mechanical Sciences*, *8*(1), 29.

Lobontiu N and Paine JS (2002) Design of circular cross-section corner-filleted flexure hinges for three-dimensional compliant mechanisms. *Journal of Mechanical Design, 124*(3), 479-484.

Lobontiu N, Cullin M, Petersen T, Alcazar JA and Noveanu S (2013) Planar compliances of symmetric notch flexure hinges: The right circularly corner-filleted parabolic design. *IEEE Transactions on Automation Science and Engineering*, *11*(1), 169-176.

Lobontiu N, Paine JS, Garcia E and Goldfarb M (2001) Corner-filleted flexure hinges *Journal of Mechanical Design*, *123*(3), 346-352.

Marković K and Zelenika S (2017) Optimized cross-spring pivot configurations with minimized parasitic shifts and stiffness variations investigated via nonlinear FEA. *Mechanics based design of structures and machines*, *45*(3), 380-394.

Smith ST, Badami VG, Dale JS and Xu Y (1997) Elliptical flexure hinges. *Review of Scientific Instruments*, *68*(3), 1474-1483.

Trease BP, Moon YM and Kota S (2005) *Journal of Mechanical Design,* 127(4): 788-798.

Weisbord L and Paros JM (1965) How to design flexure hinges. *Machine Design*, *27*(3), 151-157.

Xu W and King T (1996) Flexure hinges for piezoactuator displacement amplifiers: flexibility, accuracy, and stress considerations. *Precision engineering*, *19*(1), 4-10.